\documentclass{article} 
\usepackage{iclr2022_conference,times}

\usepackage{amsmath,amsfonts,bm}

\def\eqref#1{equation~\ref{#1}}

\def\1{\bm{1}}

\DeclareMathAlphabet{\mathsfit}{\encodingdefault}{\sfdefault}{m}{sl}
\SetMathAlphabet{\mathsfit}{bold}{\encodingdefault}{\sfdefault}{bx}{n}

\usepackage{hyperref}
\usepackage{url}
\usepackage{amsmath}
\usepackage{graphicx}
\usepackage{subcaption}
\usepackage{todonotes}
\usepackage{algorithm}
\usepackage{wrapfig}
\usepackage{algpseudocode}
\usepackage{listings}
\usepackage{algpseudocode}

\title{Goal-Guided Neural Cellular Automata: Learning to Control Self-Organising Systems}

\author{Shyam Sudhakaran, Elias Najarro \& Sebastian Risi\\
IT University of Copenhagen \\
Copenhagen, Denmark \\
\texttt{shyamsnair@protonmail.com, \{enaj,sebr\}@itu.dk} \\
}

\iclrfinalcopy 
\begin{document}

\maketitle

\begin{abstract}
Inspired by cellular growth and self-organization, Neural Cellular Automata (NCAs) have been capable of "growing" artificial cells into images, 3D structures, and even functional machines. NCAs are flexible and robust computational systems but -- similarly to many other self-organizing systems --- inherently uncontrollable during and after their growth process. We present an approach to control these type of systems called \emph{Goal-Guided Neural Cellular Automata}  (GoalNCA), which leverages goal encodings to control cell behavior dynamically at every step of cellular growth. This approach enables the NCA to continually change behavior, and in some cases, generalize its behavior to unseen scenarios. We also demonstrate the robustness of the NCA with its ability to preserve task performance, even when only a portion of cells receive goal information.


\end{abstract}

\section{Introduction}
\begin{figure}[!ht]
    \centering
    {\includegraphics[width=0.3\textwidth]{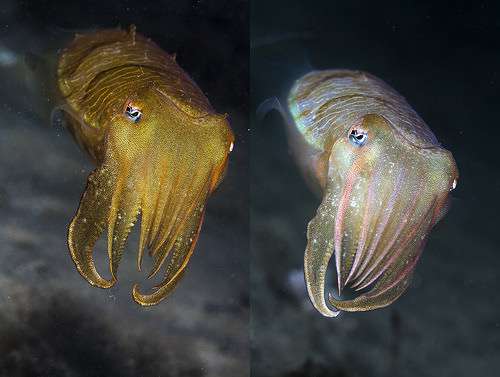} }%
    \qquad
    {\includegraphics[width=0.3\textwidth]{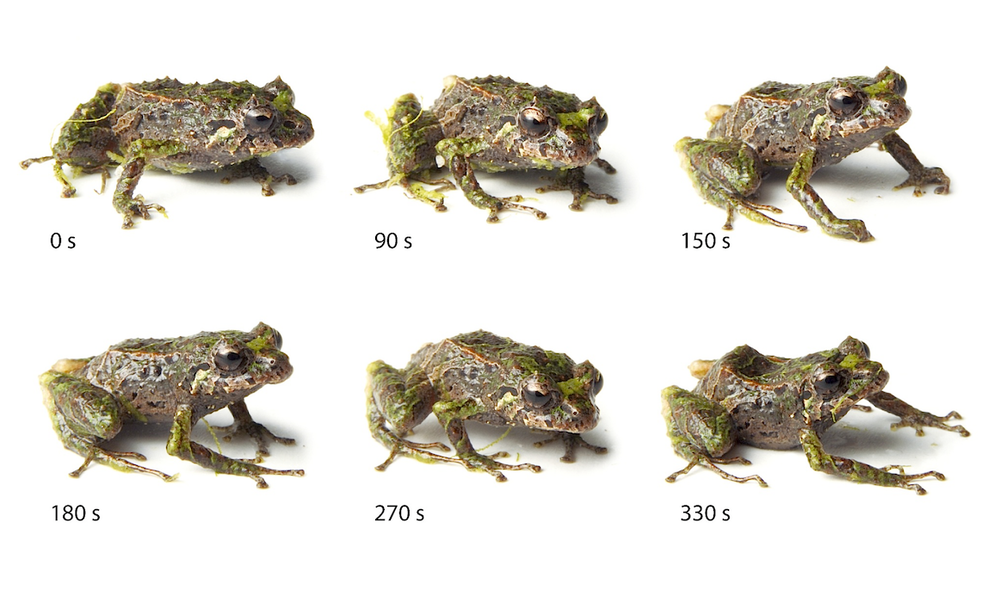} }%
    \caption{Left: Cuttlefish that can change the  color of its cells to blend in with its environment. Right:  \emph{Pristimantis mutabilis}, or mutable rainfrog, can change its skin texture from smooth to spiky. Images from \citet{morphinganimals}.}
    \label{fig:cuttlefish}
\end{figure}

Cellular growth and morphogenesis has resulted in an incredibly diverse collection of multi-cellular organisms on Earth. A single cell starts off a chain of replication and local interactions that eventually result in a fully grown organism. Some of these organisms have learned to control and manipulate characteristics of their cells dynamically, resulting in diverse behaviors, such as the ability to change skin color, texture, and even shape (Figure \ref{fig:cuttlefish}). Inspired by the flexibility and adaptability of these biological systems, we present \emph{Goal-Guided Neural Cellular Automata} (GoalNCA), an artificial self-organizing system that is able to not only grow complex shapes through the local interaction of its cells, but also guide their behaviors dynamically.

\subsection{Computational Model: Neural Cellular Automata}
Cellular Automata (CAs), inspired by biological cellular growth and self-organization, are systems that apply local rules repeatedly on artificial cells that are typically represented as a 1 or a 0 in a grid. Conway's Game of Life, a simple CA system with only four discrete local rules, is Turing Complete and  able to represent many complex behaviors. Neural Cellular Automata (NCAs) attempt to learn the local rules using compact neural networks and cell states, guided by an objective function \citep{mordvintsev2020growing,nichele2017neat,wulff1992learning}. Each cell, now represented as a vector, is composed of target channels, a living channel that determines whether a cell should be alive or dead, and hidden channels which serve as a sort of cell memory, similar to the hidden state in an RNN \citep{DBLP:journals/corr/abs-1808-03314}. This formulation has been shown to be effective at learning a variety of tasks, such as  self-classifying 2D images \citep{randazzo2020self-classifying} and physical 2D shapes \citep{walker2022physical}, game levels \citep{earle2022illuminating}, and regenerative functional 3D machines \citep{DBLP:journals/corr/abs-2103-08737}. 

Typically, an NCA starts with a small set of initial ``live'' cells, or ``seeds'', that eventually grow into a full artificial organism over time. This seed can be a static vector (with the hidden channels being set to 1), or they can be learned in order to grow a diverse set of objects from a single NCA network \citep{palm2022variational,frans2021stampca}. However, controlling a self-organizing system composed of many independent components can be challenging. As \cite{kelly2009out} writes: 

\emph{``Guiding a swarm system can only be done as a shepherd would drive a herd: by applying force at crucial leverage points, and by subverting the natural tendencies of the system to new ends.''}.

For example, NCAs can lack control over cell behaviour because  the system's stability and behaviors rely on repeated operations on initial seeds.
These cells are not trained to have dynamic behaviors during and after growth, so naively nudging a cell in a direction can cause a chain reaction that results in the collapse of the system. Our proposed variation, GoalNCA, uses encoded goal vectors to perturb and guide these cells throughout growth, giving the ability to change cell behaviors dynamically. The insights gained on NCAs (e.g.\ the system can be guided also if only a portion of the cells receive the goal information) could be useful to making self-organizing systems in general more controllable, opening interesting future research directions. 

\section{Related Work: Guiding Self-Organizing Systems}
Recently, ideas from collective intelligence and self-organizing systems have started to capture the interest of the larger machine learning community \citep{risi2021selfassemblingAI, ha2021collective}. However, self-organizing systems in general have shown to be inherently difficult to control \citep{kelly2009out,gershenson2007design}. In this section we review the two most related approaches to the one introduced in this paper, which both deal with the control of self-organization in cellular automata.

The first one is \textbf{Lenia}, an extension of classical CA, moving a discrete grid-like space to a continuous one \citep{Chan2018Dec}. A computational system classified as artificial life —or virtual creature—, Lenia exhibits a series of remarkable properties reminiscent of life forms. It thus constitutes an interesting abstract model to study the emergence of complex morphogenetic patterns \textit{in-silico}. Lenia has recently been extended for goal-guided locomotion \citep{hamon:hal-03519319}, using NCAs and curriculum learning  with sampled goals to grow creatures that can traverse through obstacles \citep{forestier2020intrinsically}. Our approach is similar in that we also grow creatures that are able to perform a variety of goals, but we explicitly incorporate goal encodings directly into the cells, and perturb the cell states to change behaviors dynamically.

The second one is \textbf{Adversarial Perturbations in Neural Cellular Automata}. The idea of changing and guiding behaviors of NCA systems has been explored before in \citet{randazzo2021adversarial}, where different methods of adversarial attacks on NCA states were evaluated. One particularly effective method was multiplying trainable symmetric perturbation matrices with NCA cell states. The authors showed that this method can change the cells' behaviors, guided by a particular target (such as changing the colors of the cells). To train these perturbation matrices, the authors first freeze a pretrained NCA's weights, and then ``fine tune'' the perturbation matrices towards goals. This setup has the advantage of training smaller parameters efficiently and re-using the existing trained NCA. In our work, instead of freezing the NCA weights and training perturbation matrices, we train a goal encoder concurrently with the NCA in order to change cell behavior throughout the growth process. We also opted for a simpler, more computationally efficient approach, where we instead add goal encoded perturbation vectors.

\section{Goal-Guided Neural Cellular Automatata}
Our system and network architecture is similar to that in \citet{mordvintsev2020growing}. The main contribution of the presented approach is a goal encoder which adds updates to every living cell's hidden states at each step in the growth cycle (Figure \ref{fig:nca_perturbation_update}). Surprisingly, this simple addition is effective in guiding cell behavior towards tasks such as morphing images and learning locomotion dynamics.

\subsection{Model Details and Training Procedure}

\begin{figure}[!ht]
    \centering
    \includegraphics[width=1\textwidth]{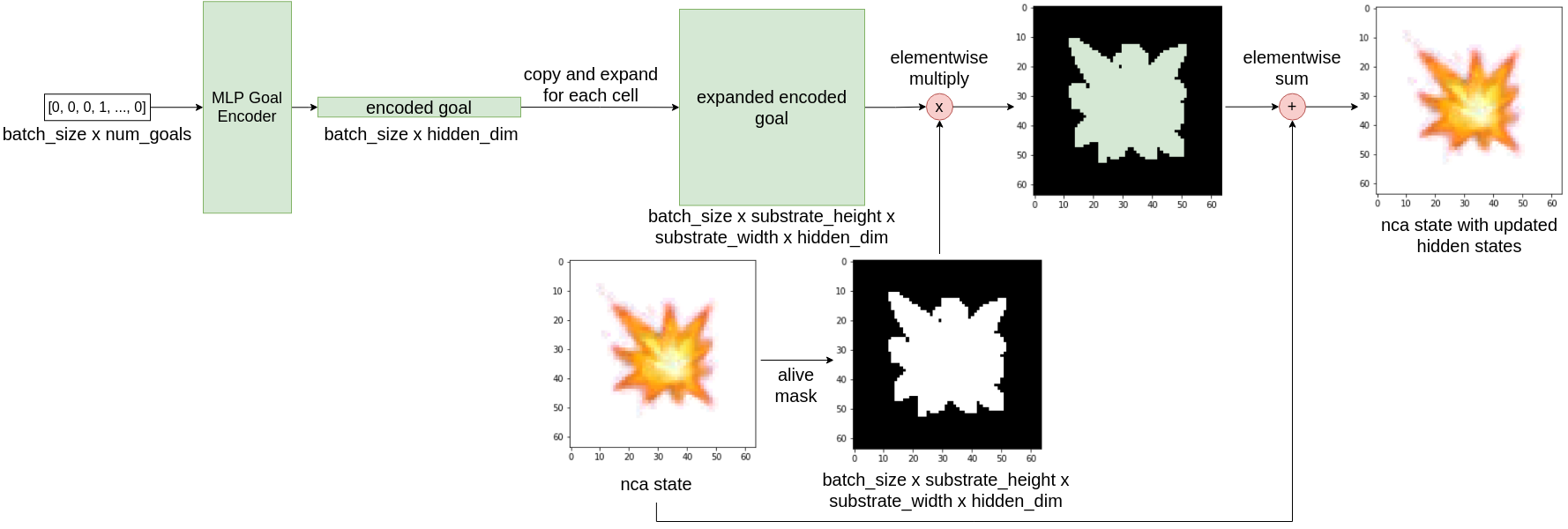}  
    \caption{How goals are encoded and added to hidden states of an NCA state. A one hot encoded goal vector is passed into an MLP goal encoder, which outputs a hidden state perturbation vector. This perturbation vector is added to every living cell.}
    \label{fig:nca_perturbation_update}
\end{figure}

\begin{figure}[!ht]
    \centering
    \includegraphics[width=1\textwidth]{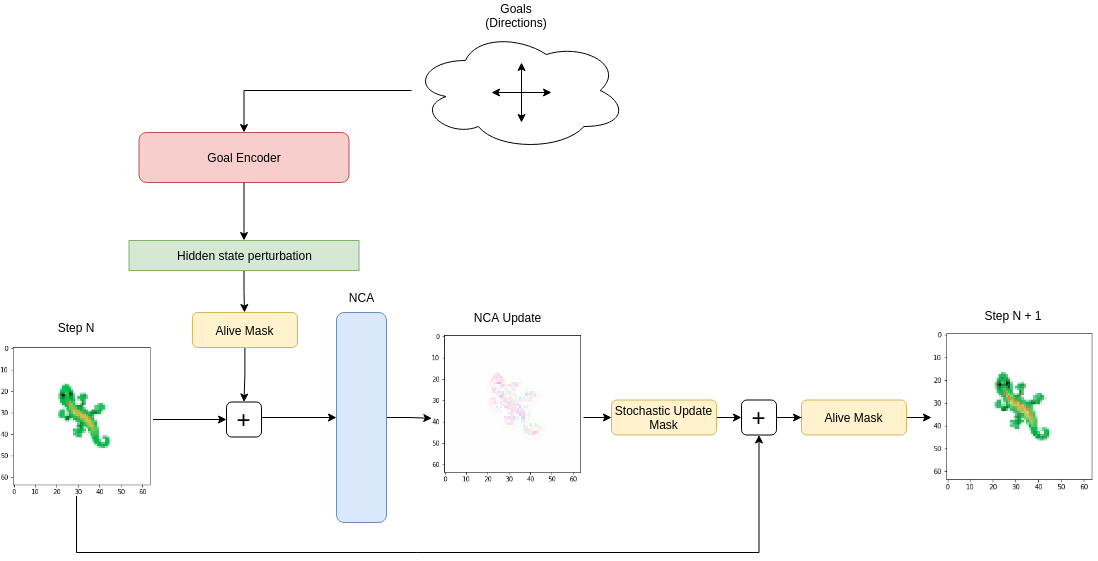}  
    \caption{Single NCA update step in the controllable locomotion experiment. A one-hot encoded direction is passed into a goal encoder to compute hidden state perturbations for live cells, which get added to previous cell states to compute the next NCA state}
    \label{fig:nca_update}
\end{figure}

Our GoalNCA model consists of two major components: A simple 3-layer MLP goal encoder and a typical NCA, similar to the one proposed in \citet{mordvintsev2020growing}. Our NCA consists of a learnable 3$\times$3 single convolution, followed by 3 1$\times$1 convolutional blocks. A single update step of our NCA is described in Figure \ref{fig:nca_update}. 

We first take the cell states at step $n$, perturb all live cells' hidden states by a encoded goal vector (Figure \ref{fig:nca_perturbation_update}), and calculate updates via forward pass through our convolutional layers. Before adding each update, we multiply it by a boolean stochastic mask, where each cell has a probability of being updated (which is referred to as $cell\_fire\_rate$). We then multiply an ``alive mask'', which utilizes the cells' living channels (3rd channel in the cell vector), where if all neighborhood cells have values $<$ 0.1, the cell is considered dead and multiplied by 0.0.
\begin{algorithm}
\caption{: GoalNCA}\label{alg:goalncasudo}
\begin{algorithmic}[ht!]
\State initialize sample pool $S$ with  $N$ initial seeds
\State initial collection of goals $G$
\State initialize NCA $f_{\theta}$ with parameters $\theta$
\State initialize goal encoder $e_{\phi}$ with parameters $\phi$
\State initialize $min\_steps$, $max\_steps$
\For{training iterations $t = 1...T$}
    \State sample NCA states from sample pool $b_{t}$
    \State sample goals and targets $g_{t}$, $target_{t}$
    \State encode $e_{t} = e_{\phi}(g_{t})$
    \State $o_{t} = b_{t}$
    \For{$num\_steps \sim [min\_steps, max\_steps]$}
        \State $o_{t} = f_{\theta}(o_{t}, e_{t})$
    \EndFor
    \State $\theta \gets \theta$ - $\nabla_{\theta}L(o_{t}, target_{t})$
    \State $\phi \gets \phi$ - $\nabla_{\phi}L(o_{t}, target_{t})$
    \State replace $b_{t}$ with $o_{t}$ in $S$
\EndFor
\end{algorithmic}
\end{algorithm}

Our training procedure, detailed in Algorithm \ref{alg:goalncasudo}, is also similar to that in \citet{mordvintsev2020growing}, with the addition of goal sampling. At the start of training, we populate the sample pool with initial seeds (single alive cell in the center with ones for its hidden values and living channel), which acts as a replay memory to ensure that the NCA is able to preserve the stability of its cells over many iterations. At every training step, we uniformly sample and encode goals and NCA states from the sample pool, replacing 2 of the states with an initial seed. The NCA states are updated for a random number of steps in the range $[min\_steps, max\_steps]$, guided by the goal encodings. Next, we update the NCA and goal encoder via MSE loss between the new NCA states' RGBA channels and the corresponding sampled goals' target images. Finally, we replace the older NCA states in the sample pool with the new NCA states.

\section{Experiments}
For all experiments, we use the Adam Optimizer \citep{kingma2017adam} with 1e-3 learning rate. We also normalize the gradients and clip NCA states between [-10, 10]. We use 64 channels for our 1$\times$1 convolutional layers and for our 3$\times$3 convolutional layer we have $num\_cell\_channels * 3$ output channels. We also trained for 100k training iterations on an NVIDIA 2080ti GPU. Additional hyperparameter details shown in Table~\ref{tab:hyperparam_details}. Code for both experiments can be found at \href{https://github.com/shyamsn97/controllable-ncas}{https://github.com/shyamsn97/controllable-ncas}

\begin{table*}[ht!]
    \centering
    \begin{tabular}{ |p{0.15\textwidth}||p{0.1\textwidth}|p{0.15\textwidth}|p{0.1\textwidth}| p{0.08\textwidth} | p{0.05\textwidth} | p{0.05\textwidth} | p{0.05\textwidth} | p{0.3\textwidth}}
         \hline
         \multicolumn{8}{|c|}{Experiment Hyperparameters} \\
         \hline
         Experiment& No. NCA params & Goal Type & Encoder & No. Hidden Cell States & Cell Fire Rate & Batch Size & Pool Size \\
         \hline
         Morphing Emoji, Image Encoder & 10,167 & one-hot vector, size 10 & Linear, 1888 params & 16 & 0.5 & 8 & 1024  \\
         \hline
         Controllable Locomotion & 10,167 & one-hot vector, size 5 & Linear, 160 params & 32 & 1.0 & 24 & 256  \\
         \hline
    \end{tabular}
    \caption{Hyperparameter Details}
    \label{tab:hyperparam_details}
\end{table*}

\subsection{Morphing Emojis}

\begin{figure}[!ht]
    \centering
    \includegraphics[width=0.65\textwidth]{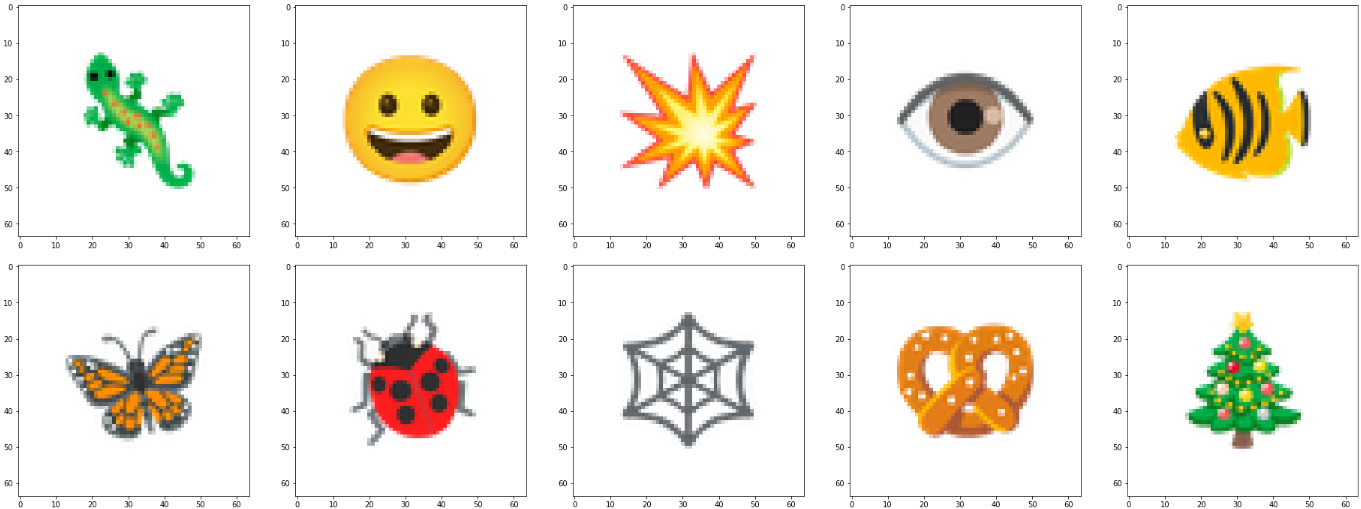}  
    \caption{Emoji dataset used in the Morphing Emojis experiment. Each image is 4$\times$64$\times$64 (RGBA channels) and is scaled between [0,1].}
    \label{fig:emojis_dataset}
\end{figure}

\begin{figure}[!ht]
    \centering
    \includegraphics[width=1\textwidth]{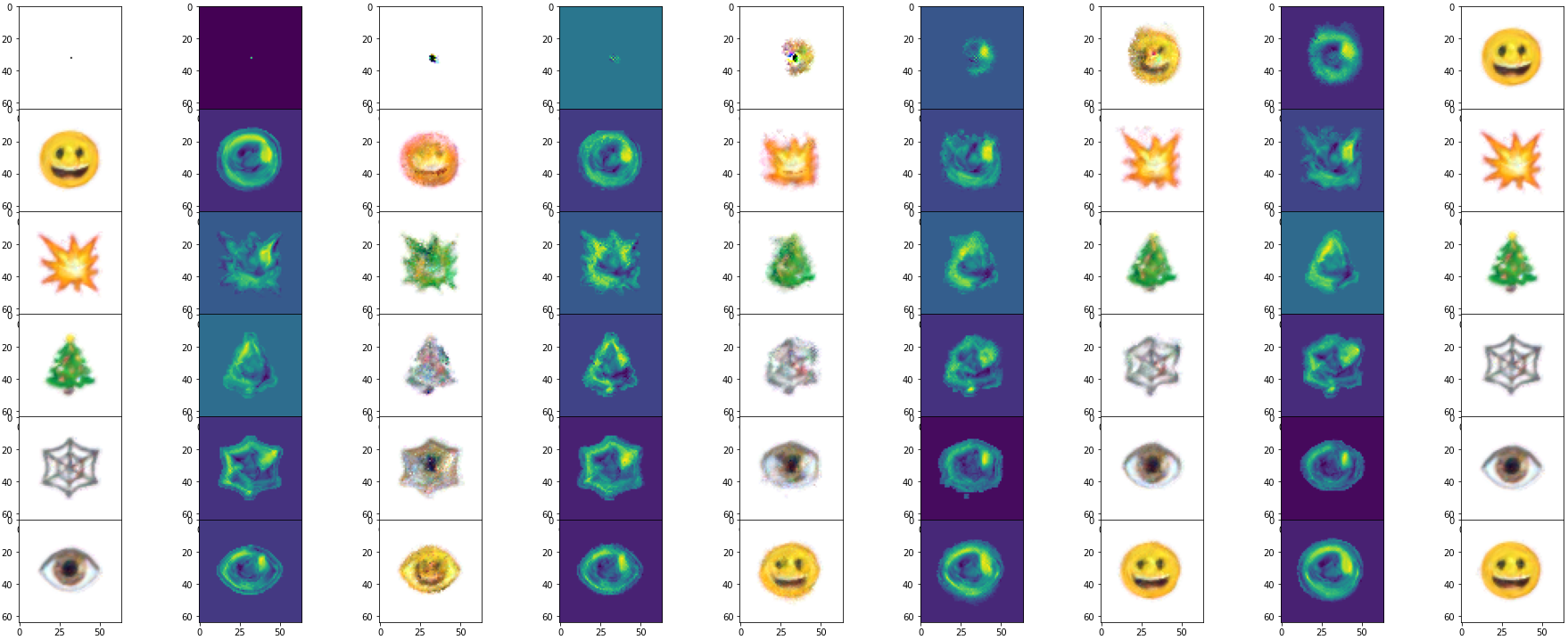}  
    \caption{Visualization of emoji morphing and hidden states over time. Emojis in column 1 are the previous rows' final grown emojis. Update steps are in intervals [8, 16, 16, 64], guided by different goal encodings corresponding to different target emojis}
    \label{fig:morphing}
\end{figure}

Inspired by organisms that can change their shape and colors, we aim to learn a GoalNCA that is able to manipulate cells to morph into different emojis, shown in Figure~\ref{fig:emojis_dataset}.

For our goal encoder, we chose a simple 3-layer MLP that takes in a one-hot encoded vector, corresponding to a particular emoji. We also experimented with a convolutional goal encoder that takes in the actual emoji, which resulted in more detailed grown emojis. However, this also introduced over 100k additional parameters.

 We train using $min\_steps = 96$ and  $max\_steps = 192$, but because of GPU limitations, we instead grow samples for a random number of steps between $[48, 96]$ twice and perform NCA updates for each. We also set ``cell fire rate'' to 0.5, denoting that a cell has a 50\% chance of receiving an update at every growth step. We found that this stabilizes training and adds robustness to the system.
 
 The GoalNCA learns to grow cells that are able to transform from any emoji in the dataset to another while keeping the cells stable, essentially learning a cell mapping between emojis. The system can also interpolate between emojis, revealing a mixture of characteristics when cells are morphing between them (Figure \ref{fig:morphing}). 
 

\subsection{Controllable Locomotion}
\begin{figure}[ht!]
    \centering
    \includegraphics[width=1\textwidth]{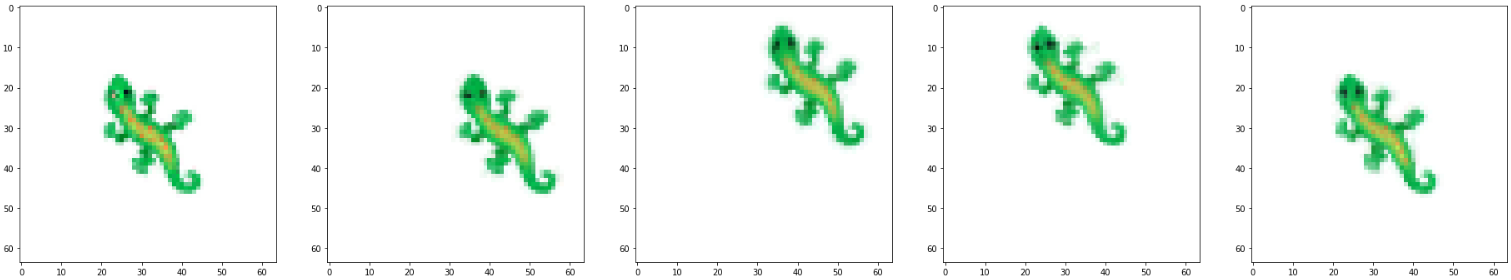}  
    \caption{Discrete Movements controlled by NCA from left to right: (1) Salamander grown for 96 steps; (2) salamander moving right for 96 steps; (3) salamander moving up for 96 steps; (4) salamander moving left for 96 steps; (5) salamander moving back down for 96 steps.}
    \label{fig:controllable_locomotion_salamander}
\end{figure}

Cellular Automata systems, such as Lenia~\citep{hamon:hal-03519319,Chan2018Dec} and Game of Life~\citep{conway1970game}, can create diverse artificial creatures that can move while also keeping consistent shape. In this experiment, we aim to extend this ability by learning a system that can not only grow and move artificial creatures, but control and learn general locomotion dynamics.

 We choose the salamander emoji as our target and use 5 different directions (stay in place, up,  down,  left, right) as our goals. We represent the directions as one-hot encoded vectors following a linear goal encoder. In our implementation, we use the Pytorch Embeddings module (\citet{paszke2019pytorch}) for simplicity. The target for each direction is the salamander shifted by $num\_steps / 8$ number of pixels in the particular direction, meaning that 8 NCA growth steps should result in a shift of one pixel. In order to achieve smooth movement, at every training step we sample random directions and use a series of 4 targets that are gradually farther away. This setup allows for the NCA to learn natural movement while also being able to retain the salamander's shape (Figure \ref{fig:controllable_locomotion_salamander}). 

\begin{figure}[!ht]
    \centering
    \includegraphics[width=0.8\textwidth]{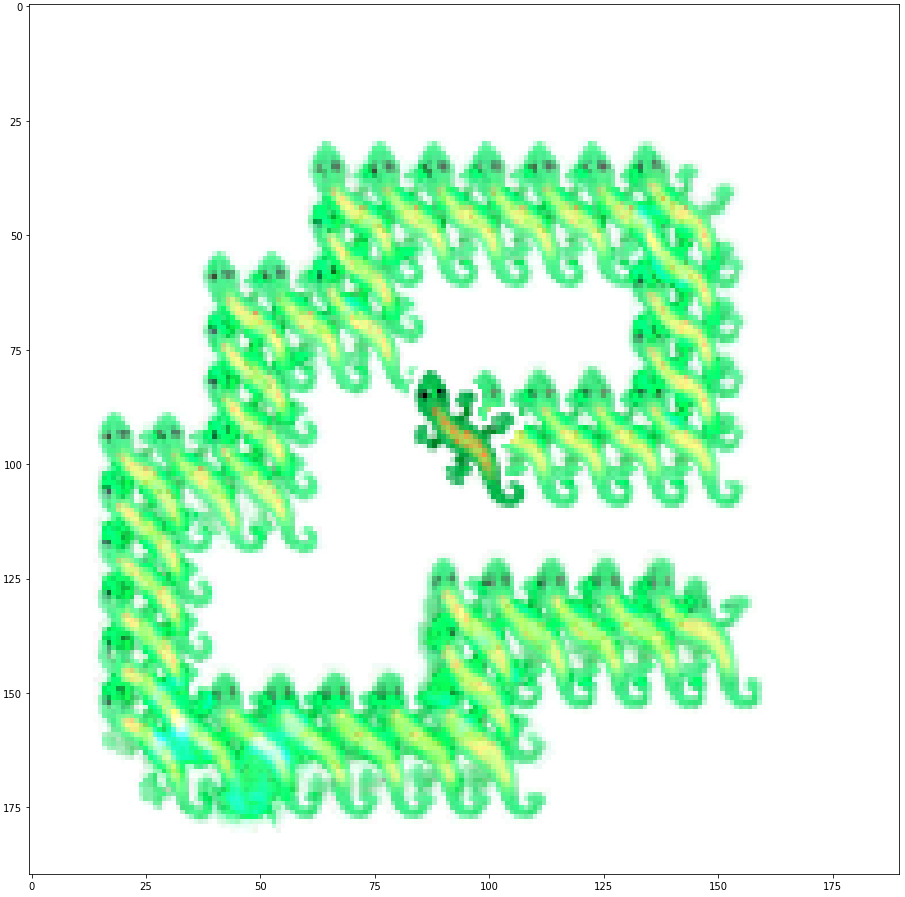}  
    \caption{Example unseen trajectory in a larger grid than what was used in training (this one is 192$\times$192 while during training the grid was only 64$\times$64). Our NCA grows a salamander in the center, and moves in a trajectory that has not been seen during training, following human input. The cells are able to constantly switch directions while keeping their consistency for thousands of steps.
    }
    \label{fig:diverse_trajectories}
\end{figure}

The GoalNCA learns the fundamentals of movement and is able to guide the cells to move freely in diverse trajectories that had not been observed during training (Figure \ref{fig:diverse_trajectories}). We observe that it is able to do this remarkably well, as it's able to move in an erratic fashion for 1000s of steps without collapsing the system. Also, because NCAs operate solely on local information, we can expand the grid size infinitely and thus discover an infinite number of trajectories. 

\subsection{Robustness to Goal Dropout}

\begin{figure}[!ht]
    \centering
    \includegraphics[width=1\textwidth]{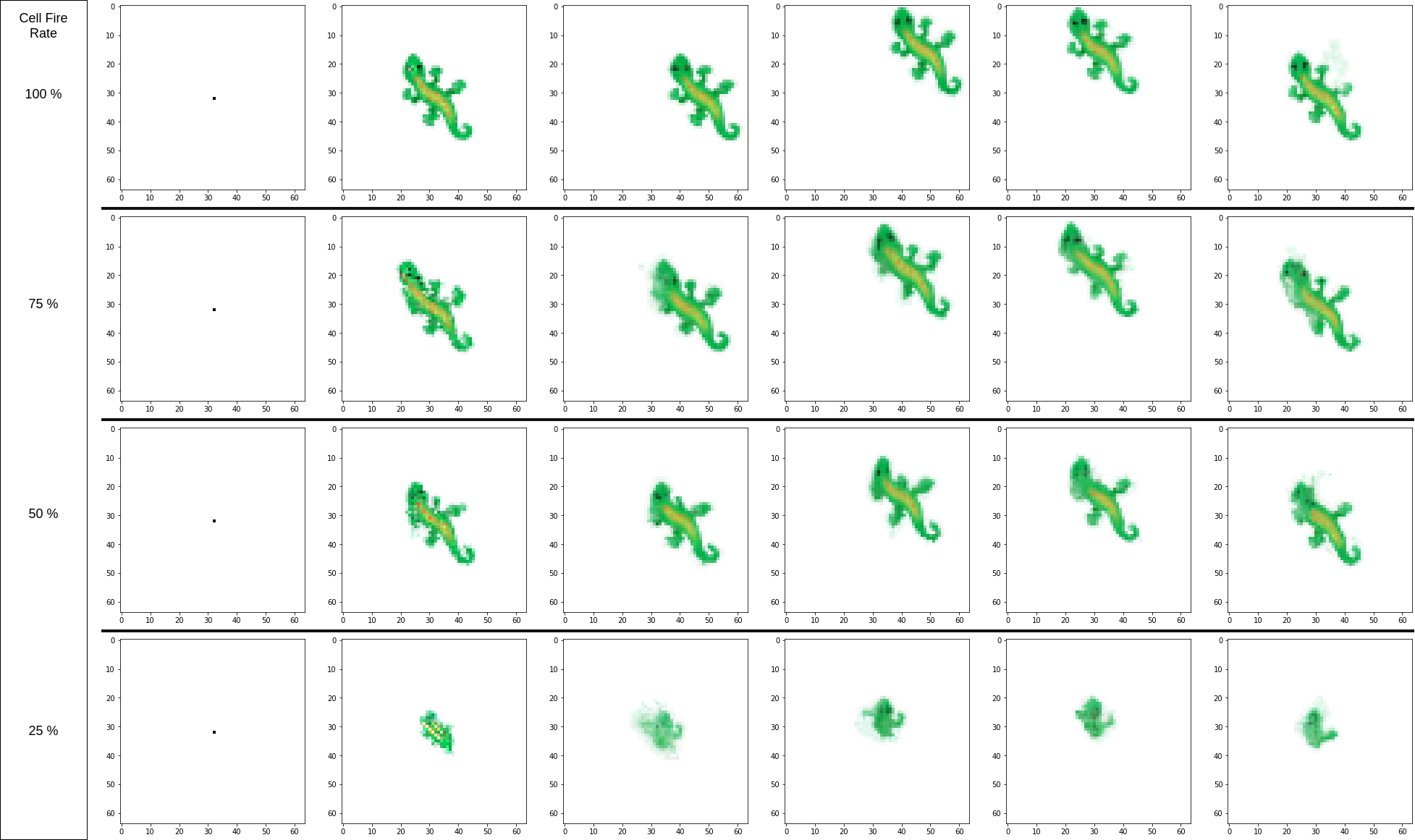}  
    \caption{Movements compared across different cell fire rates. Note that cell fire rate during training is set to 100\%.  Columns: (1) initial seed with only one live cell in the center. (2) Grown for 96 steps, guided by center goal embedding (goal is to grow and keep the salamander stationary). (3) Guided by right direction goal for 128 steps, and then guided by center goal for 96 steps. (4) Guided by up direction goal for 128 steps, and then guided by center goal for 96 steps. (5) Guided by left direction goal for 128 steps, and then guided by center goal for 96 steps. (6) Guided by down direction goal for 128 steps, and then guided by center goal for 96 steps. At this point the salamander should be back in the center of the grid.}
    \label{fig:locomotion_compare_cell_fire_rates}
\end{figure}

Self-organizing systems, leveraging local communication, are often inherently robust to adversarial modifications, such as observation dropout \citep{tang2021sensory} and damage \citep{mordvintsev2020growing,palm2022variational}. We find that in our controllable locomotion experiment, in which all cells received updates during training (cell fire rate = 1.0), the system is robust even if only a portion of cells receive an update during evaluation (Figure \ref{fig:locomotion_compare_cell_fire_rates}). In the controllable locomotion experiment, we found that if cells only have a 75\% chance of receiving updates per timestep, the grown salamander is still able to move and retain its rough shape. At 25\% - 50\%, even though the salamander shape isn't completely preserved, the cells still manage to move slightly in the goal directions. This strengthens the case for developing dynamic behaviors in self-organizing systems, which can enable them to thrive in imperfect environments with missing information.


\section{Discussion and Future work}

\begin{figure}[ht!]
    \centering
    {\includegraphics[width=0.4\textwidth]{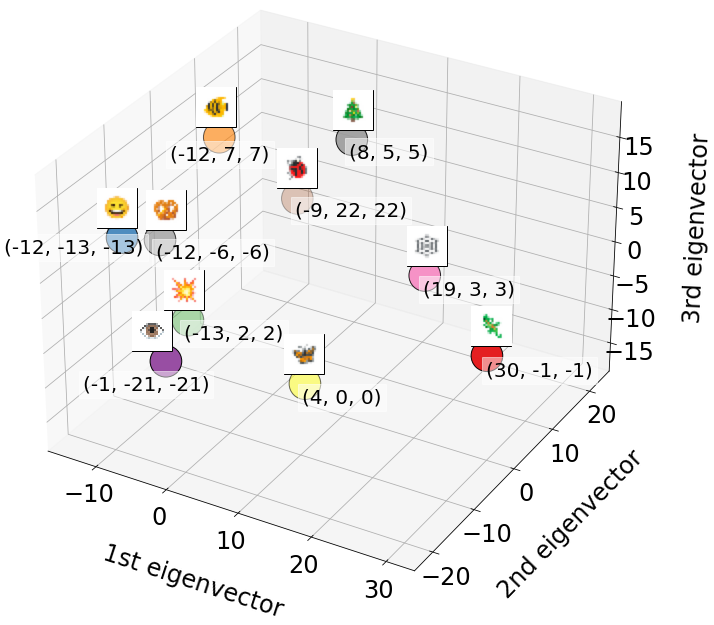} }%
    \qquad
    {\includegraphics[width=0.4\textwidth]{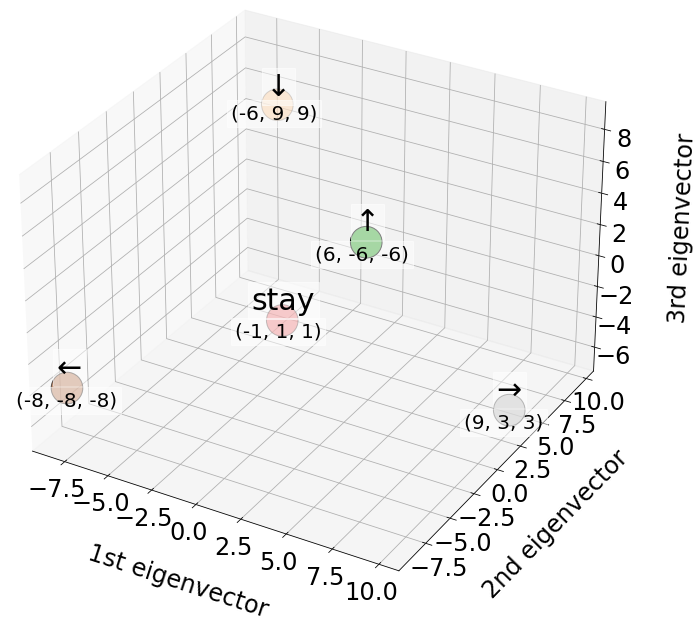} }%
    \caption{Left: PCA embedded morphing image encodings. Similar emojis are grouped together (smiley and pretzel), while those with distinct shapes are farther away from the rest (spider web, salamander). Right: PCA embedded directions encodings. Each direction has distinct behaviors associated with it, so they are far apart from each other.}
    \label{fig:embeddings}
\end{figure}

NCAs and self-organizing systems are incredibly powerful and can be robust to damage, but they lack flexibility and control over their cells after the growth process is complete. 
Here we show that by simply adding goal encoded perturbations to hidden cell states, our GoalNCA approach is able to guide cells towards diverse and robust behaviors throughout the growth cycle. We also notice that, as a byproduct of the training procedure, the GoalNCA ends up learning a goal embedding space, able to group similar goals closer together and keep others farther apart (Figure \ref{fig:embeddings}).

The GoalNCA was able to effectively guide cells towards desired behaviors while only requiring an additional small linear goal encoder ($<$ 2k params) for both experiments. In the morphing image experiment, we also tested a convolutional encoder with over 100k params, which was able to grow and morph sharper images with the tradeoff of a large increase in parameters. One limitation with the goal encoders presented is that they are not generative models, in that we cannot learn a goal distribution. This approach can be incredibly powerful, as we would be able to sample goals and diverse behaviors. This would be similar to the Variational Neural Cellular Automata (VNCA) introduced in \citet{palm2022variational}, which samples seed states in order to grow diverse images.

Our goal sampling method, essential in order for the GoalNCA to learn behaviors associated with goals, is naive and can be optimized. At every training step, we sample goals uniformly with replacement. Because we used a small goal space, our system was exposed to enough samples of each goal during training. However, in cases of large goal spaces, it is not guaranteed that goals will be observed equally and some may even be overlooked entirely. In the future, we hope to improve our goal sampling methods with ideas from curriculum learning. One possibility could be Intrinsically Motivated Goal Exploration Processes (IMGEP)  \citep{forestier2020intrinsically}, where goals are dynamically chosen based on an exploration / exploitation procedure, allowing for efficient exploration and learning with large goal spaces.

In the future, we plan to extend our approach to more complex tasks that require incorporating dynamic feedback in self-organizing systems. We are particularly interested in guiding cells that actually react to their environment by leveraging goals that represent environmental states. This could prove useful when designing self-organizing controllers, which need to be able to incorporate environment information.

\section*{Acknowledgements}
This project was supported by a Sapere Aude: DFF-Starting Grant (9063-00046B).

\bibliography{iclr2022_conference}
\bibliographystyle{iclr2022_conference}


\end{document}